\documentclass[runningheads]{llncs}
\usepackage[T1]{fontenc}
\usepackage{graphicx}
\usepackage{booktabs}
\usepackage{fontawesome5}
\usepackage{multirow}
\usepackage{amsmath}
\usepackage{amssymb}
\usepackage{bm}
\usepackage{subfigure}

\usepackage[table]{xcolor}
\definecolor{bgcell}{HTML}{EEEEEE}

\usepackage[colorlinks=true,
            linkcolor=blue,     
            citecolor=blue,     
            urlcolor=blue
           ]{hyperref}

\begin{document}

\title{\texttt{MaRS}: Robust Out-of-Distribution Detection via \underline{Ma}halanobis \underline{R}esidual \underline{S}coring}

\titlerunning{\texttt{MaRS}: Mahalanobis Residual Scoring}
\author{Francesco Di Salvo \and Sebastian Doerrich \and Christian Ledig}
\authorrunning{F. Di Salvo et al.}
%
\institute{xAILab Bamberg, University of Bamberg, Germany \\
\email{francesco.di-salvo@uni-bamberg.de}\\
}
\maketitle              
\begin{abstract}
Foundation models provide highly descriptive representations for medical images, yet their reliability degrades under distribution shifts arising from changes in patients, devices, or acquisition conditions. Reliable out-of-distribution (OOD) detection is therefore essential for safe deployment. Recent post-hoc detectors efficiently exploit frozen embeddings (\emph{e.g.}, kNN), whereas reconstruction-based OOD detection in latent feature space has seen limited adoption due to inconsistent performance. In this work, we show that the limitation of reconstruction-based methods in latent space does not stem from poor reconstruction quality, but from how reconstruction errors are scored. Standard $L_2$ residual norms collapse the anisotropic residual structure, thereby suppressing informative deviations. To address this limitation, we introduce \texttt{MaRS} (Mahalanobis Residual Scoring), a label-free OOD detector that learns an in-distribution manifold using a lightweight autoencoder and measures deviation via a Mahalanobis distance on reconstruction residuals, yielding variance-aware OOD scores. Across three imaging modalities, multiple types of distribution shift, and different model families and scales, \texttt{MaRS} outperforms established confidence-, distance-, and reconstruction-based baselines, while remaining fully post-hoc and lightweight. The code is available at   
\href{https://github.com/francescodisalvo05/mars}{github.com/francescodisalvo05/mars}. 

\end{abstract}

\section{Introduction}

Deep learning has achieved remarkable performance across a wide range of medical imaging tasks. However, despite this progress, models remain vulnerable to distribution shifts caused by changes in patient populations, imaging devices, acquisition protocols, or artifacts \cite{ktena2024generative,yoon2024domain}. Such shifts can silently degrade performance and compromise clinical reliability. Out-of-distribution (OOD) detection \cite{yang2022openood,hong2024out} therefore plays a central role in ensuring reliable and safe use of AI systems in medicine. Existing OOD detection methods differ primarily in how deviation from the in-distribution is measured. \emph{Confidence-based} methods \cite{hendrycks2017a,liang2018enhancing} use classifier uncertainty as an OOD signal, but can be fragile in medical settings, where class imbalance, weak supervision, and miscalibration can reduce the reliability of classifier confidences. \emph{Distance-based} methods, including Deep kNN \cite{sun2022out} and Mahalanobis-based approaches \cite{lee2018simple,anthony2023use,mueller2025mahalanobis}, measure deviation in feature space and benefit from the rise of foundation models \cite{dosovitskiy2020vit,oquabdinov2,simeoni2025dinov3,xu2024whole,koch2024dinobloom}. However, their performance can degrade in high-dimensional embeddings, where nearest-neighbor distances become less informative and covariance estimates become harder to regularize reliably. \emph{Reconstruction-} and \emph{subspace-based} methods offer a label-free alternative by learning structure from in-distribution features and flagging samples with large reconstruction or projection error \cite{lu2018anomaly,li2020out}. The Residual baseline \cite{ndiour2020out}, for instance, models OOD as deviation from an in-distribution subspace (via PCA or kernel PCA), while ViM \cite{wang2022vim} combines such linear residuals with classifier logits and thus relies on a well-defined classification head. Critically, Residual PCA \cite{ndiour2020out} aggregates residuals with a uniform $L_2$ norm, implicitly assuming isotropic residual structure. In latent space, however, residuals can be strongly anisotropic. Consequently, ignoring their covariance structure obscures the distinction between expected reconstruction noise and OOD deviations. 

In this work, we show that a key limitation of reconstruction- and residual-based OOD detection lies not in reconstruction quality, but in how reconstruction residuals are scored. We introduce \texttt{MaRS} (\emph{Mahalanobis Residual Scoring}), a lightweight and fully post-hoc OOD detector that applies a covariance-aware Mahalanobis distance to autoencoder reconstruction residuals. By explicitly modeling anisotropy in the residual distribution, \texttt{MaRS} avoids reliance on class labels, classifier heads, or predefined subspace assumptions, while generalizing prior residual-based approaches. Taking advantage of the strong representational power of domain-agnostic foundation models \cite{schulthess2025anomaly,liu2025does}, \texttt{MaRS} operates directly on frozen backbone features, ensuring the method is both computationally efficient and adaptable to various imaging modalities. This design further eliminates any reliance on labeled data for backbone fine-tuning, thereby enabling a label-free OOD detection framework.

\begin{figure*}[htpb]
	\centering
	\includegraphics[width=0.85\textwidth]{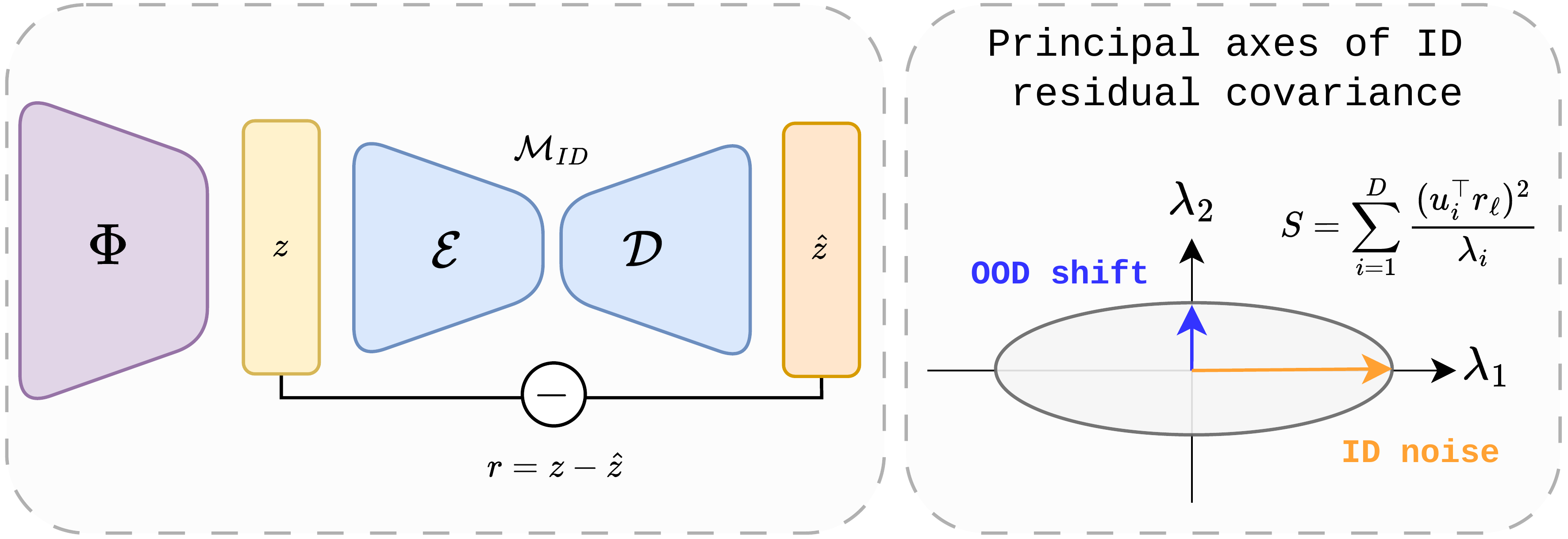}
	\caption{\label{fig:figure01}\textbf{Left}: Given a frozen backbone $\Phi$ producing features $z$, an autoencoder $(\mathcal{E}, \mathcal{D})$ learns a projection onto the in-distribution (ID) manifold $\mathcal{M}_{\mathrm{ID}}$, and the reconstruction residual $r = z - \hat z$ captures deviations from it. \textbf{Right}: Principal components of the ID residual covariance reveal that out-of-distribution (OOD) deviations concentrate in low-variance directions, which are amplified by variance-aware Mahalanobis scoring.}
\end{figure*}

Extensive experiments across three imaging modalities, multiple types of distribution shift, two backbone families, and different model scales demonstrate that \texttt{MaRS} consistently outperforms confidence-, distance-, and reconstruction-based baselines. Our results highlight the effectiveness of variance-aware residual scoring as a simple yet powerful principle for post-hoc OOD detection on frozen representations. Our contributions are as follows:

\begin{itemize}
    \item We introduce \texttt{MaRS} (see Figure~\ref{fig:figure01}), a plug-and-play, label-free OOD detection method that applies Mahalanobis distance to autoencoder reconstruction residuals, consistently outperforming strong baselines across modalities, shift types, backbone families, and model sizes.
    \item We identify residual anisotropy as a key but overlooked signal for OOD detection, and show that variance-aware Mahalanobis scoring in residual space substantially outperforms uniform $L_2$ aggregation, supported by targeted spectral analyses and ablations.
    \item We demonstrate that \texttt{MaRS} remains robust across backbone architectures and scales, and benefits from operating on pre-normalization features, where the intrinsic anisotropic variance structure exploited by residual-space Mahalanobis scoring is preserved.
\end{itemize}

\section{Method}

Foundation models provide semantically rich representations but remain vulnerable to distribution shifts~\cite{teterwak2024large}. We consider the setting of label-free, post-hoc OOD detection on frozen features and introduce \texttt{MaRS}, a lightweight adapter that estimates deviation from the in-distribution (ID) feature manifold.

\subsection{Autoencoder as a projection onto the ID manifold}

Given a frozen backbone $\Phi$, \texttt{MaRS} operates on features extracted from a fixed layer. Let $z = \Phi(x) \in \mathbb{R}^{D}$ denote the feature representation of an input $x$, where $D$ is the latent dimension. An autoencoder $(\mathcal{E}, \mathcal{D})$ produces a reconstruction $\hat z = \mathcal{D}(\mathcal{E}(z))$ and is trained on ID features by minimizing
\begin{equation}
	\mathcal{L}_{\text{AE}} = \lVert z - \hat z \rVert_2^2.
\end{equation}

Geometrically, the autoencoder implements a data-driven projection
\begin{equation}
	\Pi_{\mathcal{M}_{\mathrm{ID}}}(z) = \mathcal{D}(\mathcal{E}(z)) = \hat z
\end{equation}
onto the ID manifold $\mathcal{M}_{\mathrm{ID}}$. ID samples lie close to this manifold and are reconstructed accurately, whereas OOD samples are mapped back toward $\mathcal{M}_{\mathrm{ID}}$, inducing structured reconstruction errors that reflect deviation from the learned manifold.

\subsection{Residual space as a drift descriptor}

We define the reconstruction residual as $r(x) = z - \hat z \in \mathbb{R}^{D}$. Intuitively, $r(x)$ measures how much the feature representation must be ``pulled back'' to lie on the ID manifold. For ID samples, residuals are small and primarily reflect reconstruction imperfections. For OOD samples, residuals exhibit systematic components that are rare or absent under the ID residual distribution, encoding informative drift signals.

\subsection{Mahalanobis scoring on residuals}

A standard reconstruction score utilizes a uniform $L_2$ norm, $\|r\|_2^2 = \sum_{i=1}^{D} r_i^2$,
which treats all residual directions equally. This is suboptimal when ID residuals are anisotropic: some directions correspond to high-variance reconstruction noise, while others are highly stable and thus more sensitive to distributional shifts. We capture this structure by estimating the ID residual covariance
\begin{equation}
\Sigma = \mathrm{Cov}\!\left(r(x)\,|\,x\sim \mathrm{ID}\right),
\end{equation}
and by defining the \texttt{MaRS} score as a Mahalanobis distance in residual space:
\begin{equation}
S_{\texttt{MaRS}}(x) = r(x)^\top \Sigma^{-1} r(x).
\end{equation}

Let $\Sigma = U \Lambda U^\top$ be the eigendecomposition of the residual covariance, where $U = [u_1,\dots,u_D]$ contains the principal directions and $\Lambda = \mathrm{diag}(\lambda_1,\dots,\lambda_D)$ the corresponding eigenvalues. The score can then be written as

\begin{equation}
S_{\texttt{MaRS}}(x)
= \sum_{i=1}^{D} \frac{\big(u_{i}^\top r(x)\big)^2}{\lambda_{i}},
\end{equation}
which amplifies deviations along low-variance (stable) residual directions via $\lambda_{i}^{-1}$, while high-variance directions are downweighted. 

Unlike Mahalanobis scoring in embedding space based on class-conditional means \cite{mueller2025mahalanobis}, our formulation is label-free and does not assume well-separated class clusters.  In medical image analysis, class boundaries are often ambiguous, and labels can be scarce or noisy. Subtracting the learned projection, ${r = z - \Pi_{\mathcal{M}_{\mathrm{ID}}}(z)}$, suppresses class-semantic variation and isolates deviations orthogonal to the ID manifold, making a single global covariance $\Sigma$ sufficient. OOD detection is performed by thresholding $S_{\texttt{MaRS}}(x)$ using a percentile-based criterion on ID validation data (\emph{e.g.}, at the $95^{\text{th}}$ percentile).

\section{Experimental results}

\subsection{OOD Detection}

\noindent \textbf{Methods} \, We compare \texttt{MaRS} against a diverse set of established baselines. Confidence-based methods include MSP \cite{hendrycks2017a}, ODIN \cite{liang2018enhancing}, and ViM \cite{wang2022vim}. Distance-based baselines include Deep kNN \cite{sun2022out}, one-class SVM (OCSVM) \cite{scholkopf1999support}, and Mahalanobis++ (Maha) \cite{mueller2025mahalanobis}, all operating directly in latent feature space. Notably, the latter exhibits recent state-of-the-art performance \cite{mueller2025mahalanobis}, relying on class-conditional Gaussian modeling and feature normalization. We further include \textit{Residual}, which measures the deviation from a low-variance linear subspace estimated via PCA, following \cite{ndiour2020out}. Finally, we evaluate reconstruction-based methods using a deterministic autoencoder (AE), where OOD scores are computed as the $L_2$ norm of reconstruction residuals \cite{ndiour2020out}. \texttt{MaRS} builds upon this reconstruction framework by replacing the uniform $L_2$ with a variance-aware Mahalanobis scoring of residuals.

\noindent\textbf{Datasets} \, We evaluate \texttt{MaRS} on multiple medical imaging datasets covering diverse modalities and shift types. First, we include the established MIDOG histopathology benchmark~\cite{aubreville2023comprehensive} (CC BY 4.0), using the splits of OpenMIBOOD \cite{gutbrod2025openmibood}. ID data consist of mitotic and non-mitotic cell crops extracted from H\&E-stained whole-slide images. Covariate shifts (CS) arise from changes in imaging devices, while near-OOD datasets introduce semantic shifts across cancer types, species, and acquisition settings. Far-OOD samples originate from unrelated medical applications, including cervical cancer cell images~\cite{amorim2020novel} (Apache 2.0) and breast FNAC cytology~\cite{saikia2019comparative} (license not disclosed). We further consider two additional imaging modalities. For chest X-ray imaging, adult chest X-rays~\cite{Wang2017ChestXRay8HC} are treated as ID data, while pediatric chest X-rays~\cite{Kermany2018} serve as covariate-shifted data. For dermatoscopy, we use the HAM10000 dataset~\cite{tschandl2018ham10000}, treating non-melanocytic skin lesions as in-distribution (ID) data and melanocytic lesions as out-of-distribution (OOD) samples. This choice induces a low-data, imbalanced ID regime, with training restricted to a small and heterogeneous lesion set ($N{=}1{,}535$). For dermatoscopy (CC BY-NC 4.0) and chest X-ray (CC BY 4.0), we adopt the data collection of Yang et al.~\cite{yang2023medmnist}, using the training split as ID data and the corresponding test splits for evaluation. Due to the limited size of the pediatric chest X-ray test set, its training split is used as OOD data.

\noindent \textbf{Implementation details} \,
We adopt off-the-shelf backbones and avoid any reliance on labeled data for backbone fine-tuning.  All methods operate on last-layer embeddings extracted from a supervised ViT/S backbone~\cite{dosovitskiy2020vit}. MSP, ODIN, and ViM train a lightweight linear head for $50$ epochs using SGD with a learning rate of $10^{-3}$. ODIN and Deep kNN use the default hyperparameters of~\cite{zhang2024openood}. We additionally include Mahalanobis++~\cite{mueller2025mahalanobis}, which applies feature normalization prior to distance computation. Reconstruction-based methods (AE, \texttt{MaRS}) employ a two-layer MLP encoder and decoder (hidden ratio $0.5$, bottleneck dimension $128$) trained for $50$ epochs with SGD (learning rate $10^{-3}$). Unless stated otherwise, results are reported as mean AUROC over three runs. While latent normalization is a prerequisite for standard distance-based baselines to prevent dimensional dominance, it collapses the anisotropic variance that characterizes meaningful OOD signals. \texttt{MaRS} preserves this signal by operating on pre-normalization features, using the Mahalanobis metric to statistically whiten the residual space.

\begin{table}[h!]
	\centering
	\footnotesize
	\caption{\label{tab:table1}OOD detection performance measured by average AUROC ($\uparrow$). MIDOG results are reported under covariate (CS), near-OOD, and far-OOD shifts. Chest X-ray and dermatoscopy correspond to covariate and semantic shifts, respectively. The Average AUROC ($\uparrow$) column reports the mean AUROC, and the Average FPR@95 ($\downarrow$) column reports the mean false positive rate at 95\% true positive rate, both averaged across all benchmarks. Results are further averaged over three seed runs, and the \textbf{best two methods} per column are bolded. Overall, \texttt{MaRS} significantly outperforms Mahalanobis++ ($\dagger$, paired Wilcoxon signed-rank test, $p < 0.05$).}
	\begin{tabular}{clccccccccc}
\toprule
\textit{Unsup} &
& \multicolumn{3}{c}{\textbf{MIDOG}}
& \textbf{X-Ray}
& \textbf{Derma}
& \multicolumn{2}{c}{\textbf{Average}}\\
\cmidrule(lr){3-5}
\cmidrule(lr){6-6}
\cmidrule(lr){7-7}
\cmidrule(lr){8-9}

& 
& \textit{CS}
& \textit{near-OOD}
& \textit{far-OOD}
& \textit{CS}
& \textit{Semantic}
& AUROC $\uparrow$
& FPR@95 $\downarrow$ \\
\midrule

$\times$ & MSP & $46.71$ & $48.93$ & $62.73$ & $71.63$ & $48.99$ & $55.80$ & $88.02$ \\
$\times$ & ODIN & $45.53$ & $48.98$ & $63.44$ & $71.63$ & $55.20$ & $56.96$ & $87.33$ \\
$\times$ & ViM & $45.84$ & $47.02$ & $76.27$ & $59.26$ & $\textbf{58.43}$ & $57.36$ & $81.05$ \\
\checkmark & Residual & $92.09$ & $82.92$ & $\textbf{100.00}$ & $92.90$ & $57.25$ & $85.03$ & $45.31$ \\
\checkmark & OCSVM & $87.68$ & $81.21$ & $99.89$ & $91.80$ & $49.27$ & $81.97$ & $47.95$ \\
\checkmark & Deep kNN & $90.92$ & $80.10$ & $99.95$ & $90.55$ & $57.42$ & $83.79$ & $47.88$ \\
\checkmark & AE & $86.48$ & $77.80$ & $99.87$ & $92.73$ & $50.90$ & $81.56$ & $50.72$ \\
$\times$ & Maha & $\textbf{92.56}$ & $\textbf{83.80}$ & $\textbf{99.99}$ & $\textbf{93.09}$ & $57.24$ & $\textbf{85.34}$ & $\textbf{43.59}$ \\ \midrule
\checkmark & \texttt{MaRS} & $\textbf{93.43}$ & $\textbf{83.43}$ & $99.98$ & $\textbf{94.00}$ & $\textbf{59.24}$ & $\textbf{86.01}$$^\dagger$ & $\textbf{42.72}$$^\dagger$ \\
\bottomrule
\end{tabular}
\end{table}

\noindent \textbf{Results} \, As shown in Table \ref{tab:table1}, confidence-based detectors (MSP, ODIN, ViM) underperform on MIDOG and X-Ray, while ViM is the second-best method on dermatoscopy. Among label-free latent-space methods, Residual and Deep kNN achieve competitive performance, confirming the utility of residual-based signals. However, extending reconstruction to a nonlinear autoencoder and scoring via a uniform $L_2$ norm leads to a clear performance drop. Mahalanobis++ ranks second overall, but relies on class-conditional statistics and exhibits reduced robustness in the low-data dermatoscopy setting. In contrast, \texttt{MaRS} significantly outperforms Mahalanobis++ across 3 runs and 5 dataset splits, in both AUROC and FPR@95 (paired Wilcoxon signed-rank test, $p<0.05$), while remaining fully label-free. As Mahalanobis++ is deterministic, its results are aligned with the three-seed evaluation protocol of \texttt{MaRS} to enable paired statistical testing.

\subsection{A closer look at the residual space}

This ablation investigates \emph{why} Mahalanobis scoring in residual space outperforms a uniform $L_2$ norm. We estimate the residual covariance on the MIDOG training set, evaluate in-distribution (ID) behavior on the MIDOG test set, and use FNAC2019 as an OOD dataset. We analyze the principal components (PCs) of the ID residual covariance and measure the mean squared residual deviation $\mathbb{E}_x[(u_i^\top r)^2]$ of ID and OOD samples along each principal axis $u_i$. \newline

\begin{figure*}[htpb]
	\centering
	\includegraphics[width=0.9\textwidth]{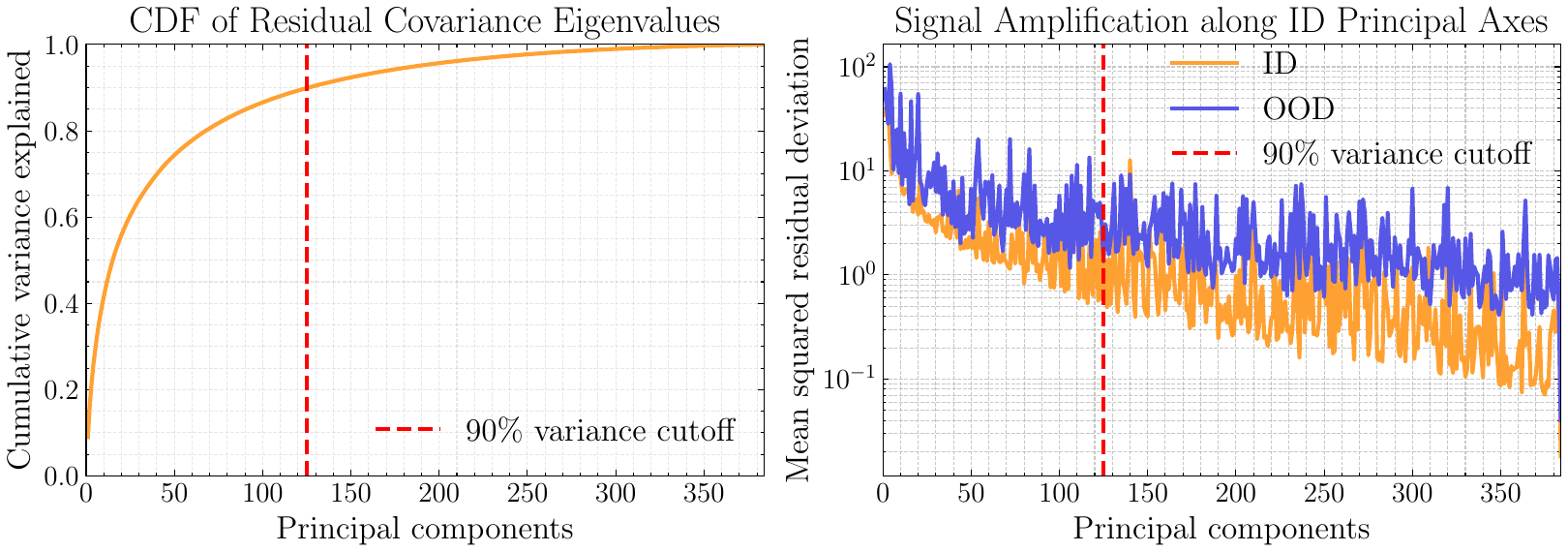}
	\caption{\label{fig:residual_space}Analysis of residual space for MIDOG+FNAC. \textbf{Left}: CDF of eigenvalues of the ID residual covariance, indicating a highly anisotropic residual distribution. \textbf{Right}: Mean squared residual deviation $\mathbb{E}_x[(u_i^\top r)^2]$ of ID and OOD samples along each principal component $u_i$. OOD residuals exhibit larger and more persistent deviations precisely in the low-variance directions, motivating variance-aware Mahalanobis scoring.} 
\end{figure*}

\noindent \textbf{Results} \, Figure~\ref{fig:residual_space} reveals two key findings. (i)~ The residual covariance is highly anisotropic: over $90\%$ of the variance is concentrated in fewer than one-third of the principal components, with the remaining components exhibiting very low variance (small eigenvalues). (ii)~When projecting ID and OOD residuals onto these axes, the relative separation between the OOD deviation (blue line) and the ID variance (orange line) is predominantly observed in the low-variance directions. Quantitatively, for each PC $u_i$, we define a normalized residual energy ratio $\rho_i = \mathbb{E}_{\text{OOD}}[(u_i^\top r)^2] / \mathbb{E}_{\text{ID}}[(u_i^\top r)^2]$. Averaging $\rho_i$ over high-variance components (explaining the first $90\%$ of ID variance) and low-variance components separately, we observe a $1.64\times$ larger ratio in the low-variance subspace, indicating that OOD deviations concentrate along stable residual directions. To exploit this property, we employ Mahalanobis scoring, which reweights each PC by the inverse of its variance ($\lambda_i^{-1}$). This ensures that OOD deviations along the low-variance directions are most heavily penalized (since $\lambda_i^{-1}$ is large), thereby maximizing the separation between the ID residual noise and the OOD signal.

\subsection{Backbone size and normalization sensitivity}

\noindent\textbf{Backbone size} \, We assess the generalizability of \texttt{MaRS} across supervised (ViT \cite{dosovitskiy2020vit}) and self-supervised (DINOv3 \cite{simeoni2025dinov3}) models at small (S) and base (B) scales. This evaluates the detector's robustness to the increased feature dimensionality ($D=384$ vs. $768$) and thereby induced data sparsity. Using Mahalanobis++ as a competitive latent-space baseline, we isolate whether scoring residuals remains superior as representational complexity grows. We focus this analysis on chest X-ray (covariate shift) and dermatoscopy (semantic shift). \newline

\begin{figure*}[htpb]
    \centering

    \begin{minipage}[t]{0.63\textwidth}
        \centering
        \includegraphics[width=\linewidth]{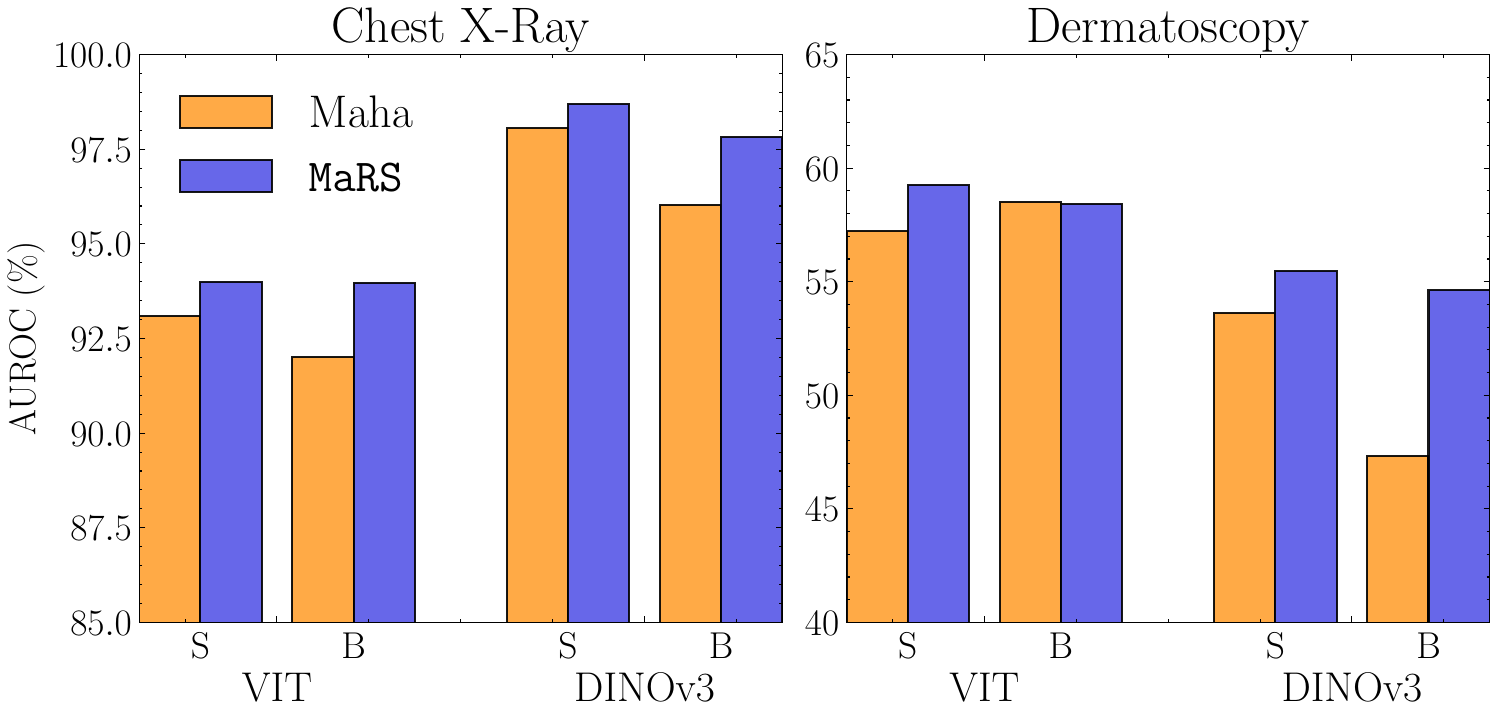}
    \end{minipage}
    \hfill
	\begin{minipage}[t]{0.32\textwidth}
        \centering
        \includegraphics[width=\linewidth]{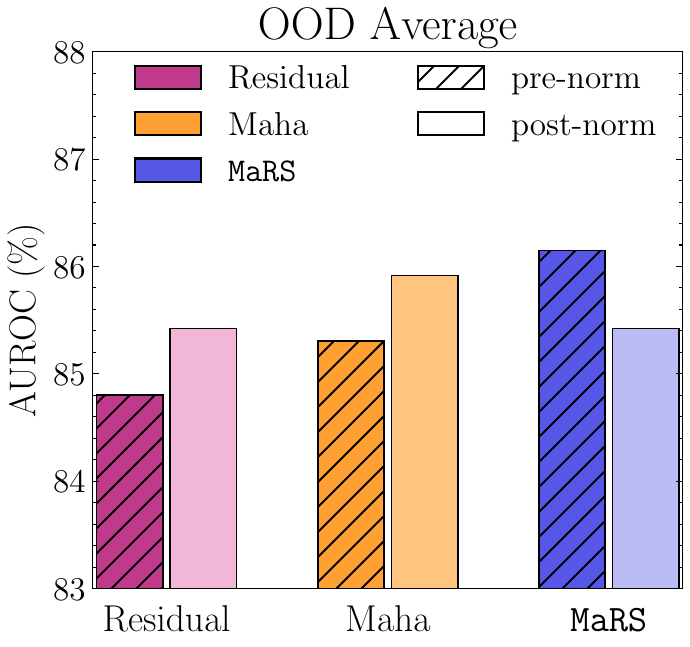}
    \end{minipage}
    \caption{
        \label{fig:backbone_tsize} \textbf{Left}: AUROC on chest X-ray and dermatoscopy for ViT and DINOv3 backbones at small (S) and base (B) scales. \texttt{MaRS} consistently outperforms Mahalanobis++ and is less sensitive to increased feature dimensionality. \textbf{Right}: Average AUROC of the top three methods under pre- and post-normalization. While normalization improves latent-space baselines, it suppresses variance anisotropy and degrades \texttt{MaRS}, confirming that residual-based Mahalanobis scoring benefits from pre-normalization features.
    }
\end{figure*}

\noindent\textbf{Results} \, As shown in Figure~\ref{fig:backbone_tsize} (left), \texttt{MaRS} outperforms Mahalanobis++ across almost all configurations. These improvements are consistent across both supervised (ViT) and self-supervised (DINOv3) models, indicating that the benefits of residual-space scoring are largely independent of the pretraining paradigm. Crucially, while Mahalanobis++ generally exhibits a notable performance decay when scaling from small to base models, likely due to the unreliable estimation of high-dimensional global covariance, \texttt{MaRS} remains remarkably robust. This suggests that reconstruction residuals act as an implicit filter, suppressing high-dimensional variability while preserving directions most sensitive to distributional shifts, consistent with the anisotropic structure of the residual space. \newline

\noindent\textbf{Pre- vs.\ post-normalization} \, We study the impact of the backbone's final LayerNorm on OOD detection performance across all datasets (average AUROC). We evaluate all methods under both \emph{pre-normalization} and \emph{post-normalization} features. Latent-space methods such as Mahalanobis++ and PCA-based residual baselines are conventionally applied to normalized features, as normalization stabilizes distances and improves numerical conditioning. In contrast, \texttt{MaRS} benefits from un-normalized features, preserving variance anisotropy. 

\noindent\textbf{Results} \, Figure~\ref{fig:backbone_tsize} (right) reports average AUROC across datasets for the top-performing methods under pre- and post-normalization. As expected, while normalization improves the performance of latent-space baselines, it degrades \texttt{MaRS}. This behavior is consistent with our analysis: normalization flattens the variance spectrum and suppresses the low-variance directions where OOD deviations are most informative. By preserving this anisotropy, \texttt{MaRS} benefits from pre-normalized features, outperforming reference methods. 

\section{Conclusions}

\textbf{Limitations} \, Although \texttt{MaRS} is highly effective for covariate and semantic shifts, commonly observed in medical image analysis, its performance depends on the quality of the underlying manifold estimation. Future work will explore whether specialized medical backbones, rather than domain-agnostic ones, further refine the residual space anisotropy. Additionally, while we focus on single-modality OOD, extending \texttt{MaRS} to multimodal data presents a promising direction. \newline

\noindent \textbf{Discussion} \, Our results show that a central limitation of existing reconstruction- and residual-based OOD detectors lies in the implicit assumption of isotropic reconstruction errors. \texttt{MaRS} addresses this limitation by replacing uniform $L_2$ aggregation with covariance-aware Mahalanobis scoring in residual space. Our analysis demonstrates that OOD deviations are most pronounced along low-variance residual directions, which are largely ignored by standard $L_2$ norms but explicitly amplified by the Mahalanobis metric. As a result, \texttt{MaRS} consistently outperforms latent-space and reconstruction-based baselines, while remaining label-free, post-hoc, and robust to high-dimensional feature representations and data sparsity.

\begin{credits}
\subsubsection{\ackname} This study was funded through the Hightech Agenda Bayern (HTA) of the Free State of Bavaria, Germany.

\subsubsection{\discintname} The authors have no competing interests to declare that are relevant to the content of this article. 
\end{credits}

%
%

\bibliographystyle{splncs04}
\bibliography{Paper-0289.bib}

@inproceedings{
mueller2025mahalanobis,
title={Mahalanobis++: Improving {OOD} Detection via Feature Normalization},
author={Maximilian M{\"u}ller and Matthias Hein},
booktitle={Forty-second International Conference on Machine Learning},
year={2025},
}

@inproceedings{sun2022out,
  title={Out-of-distribution detection with deep nearest neighbors},
  author={Sun, Yiyou and Ming, Yifei and Zhu, Xiaojin and Li, Yixuan},
  booktitle={International conference on machine learning},
  year={2022},
  organization={PMLR}
}

@inproceedings{koch2024dinobloom,
  title={DinoBloom: a foundation model for generalizable cell embeddings in hematology},
  author={Koch, Valentin and Wagner, Sophia J and Kazeminia, Salome and Sancar, Ece and Hehr, Matthias and Schnabel, Julia A and Peng, Tingying and Marr, Carsten},
  booktitle={International Conference on Medical Image Computing and Computer-Assisted Intervention},
  pages={520--530},
  year={2024},
  organization={Springer}
}

@article{xu2024whole,
  title={A whole-slide foundation model for digital pathology from real-world data},
  author={Xu, Hanwen and Usuyama, Naoto and Bagga, Jaspreet and Zhang, Sheng and Rao, Rajesh and Naumann, Tristan and Wong, Cliff and Gero, Zelalem and Gonz{\'a}lez, Javier and Gu, Yu and others},
  journal={Nature},
  volume={630},
  number={8015},
  pages={181--188},
  year={2024},
  publisher={Nature Publishing Group UK London}
}

@article{lee2018simple,
  title={A simple unified framework for detecting out-of-distribution samples and adversarial attacks},
  author={Lee, Kimin and Lee, Kibok and Lee, Honglak and Shin, Jinwoo},
  journal={NIPS},
  year={2018}
}

@inproceedings{schulthess2025anomaly,
  title={Anomaly Detection by Clustering DINO Embeddings Using a Dirichlet Process Mixture},
  author={Schulthess, Nico and Konukoglu, Ender},
  booktitle={International Conference on Medical Image Computing and Computer-Assisted Intervention},
  pages={46--56},
  year={2025},
  organization={Springer}
}

@inproceedings{anthony2023use,
  title={On the use of mahalanobis distance for out-of-distribution detection with neural networks for medical imaging},
  author={Anthony, Harry and Kamnitsas, Konstantinos},
  booktitle={International workshop on uncertainty for safe utilization of machine learning in medical imaging},
  pages={136--146},
  year={2023},
  organization={Springer}
}

@article{yang2022openood,
  title={Openood: Benchmarking generalized out-of-distribution detection},
  author={Yang, Jingkang and Wang, Pengyun and Zou, Dejian and Zhou, Zitang and Ding, Kunyuan and Peng, Wenxuan and Wang, Haoqi and Chen, Guangyao and Li, Bo and Sun, Yiyou and others},
  journal={Advances in Neural Information Processing Systems},
  volume={35},
  year={2022}
}

@article{ktena2024generative,
  title={Generative models improve fairness of medical classifiers under distribution shifts},
  author={Ktena, Ira and Wiles, Olivia and Albuquerque, Isabela and Rebuffi, Sylvestre-Alvise and Tanno, Ryutaro and Roy, Abhijit Guha and Azizi, Shekoofeh and Belgrave, Danielle and Kohli, Pushmeet and Cemgil, Taylan and others},
  journal={Nature Medicine},
  volume={30},
  number={4},
  year={2024},
  publisher={Nature Publishing Group US New York}
}

@article{yoon2024domain,
  title={Domain generalization for medical image analysis: A review},
  author={Yoon, Jee Seok and Oh, Kwanseok and Shin, Yooseung and Mazurowski, Maciej A and Suk, Heung-Il},
  journal={Proceedings of the IEEE},
  year={2024},
  publisher={IEEE}
}

@article{lu2018anomaly,
  title={Anomaly detection for skin disease images using variational autoencoder},
  author={Lu, Yuchen and Xu, Peng},
  journal={arXiv:1807.01349},
  year={2018}
}

@inproceedings{li2020out,
  title={Out-of-distribution detection for skin lesion images with deep isolation forest},
  author={Li, Xuan and Lu, Yuchen and Desrosiers, Christian and Liu, Xue},
  booktitle={International Workshop on Machine Learning in Medical Imaging},
  pages={91--100},
  year={2020},
  organization={Springer}
}

@article{hong2024out,
  title={Out-of-distribution detection in medical image analysis: A survey},
  author={Hong, Zesheng and Yue, Yubiao and Chen, Yubin and Cong, Lele and Lin, Huanjie and Luo, Yuanmei and Wang, Mini Han and Wang, Weidong and Xu, Jialong and Yang, Xiaoqi and others},
  journal={arXiv:2404.18279},
  year={2024}
}

@article{Kermany2018,
   author = {Daniel S. Kermany and Michael Goldbaum and Wenjia Cai and Carolina C.S. Valentim and Huiying Liang and Sally L. Baxter and Alex McKeown and Ge Yang and Xiaokang Wu and Fangbing Yan and Justin Dong and Made K. Prasadha and Jacqueline Pei and Magdalena Ting and Jie Zhu and Christina Li and Sierra Hewett and Jason Dong and Ian Ziyar and Alexander Shi and others},
   issn = {0092-8674},
   issue = {5},
   journal = {Cell},
   pages = {1122-1131.e9},
   pmid = {29474911},
   publisher = {Cell Press},
   title = {Identifying Medical Diagnoses and Treatable Diseases by Image-Based Deep Learning},
   volume = {172},
   year = {2018},
}

@inproceedings{liang2018enhancing,
  title={Enhancing The Reliability of Out-of-distribution Image Detection in Neural Networks},
  author={Liang, Shiyu and Li, Yixuan and Srikant, R},
  booktitle={International Conference on Learning Representations},
  year={2018}
}

@article{yang2023medmnist,
  title={Medmnist v2-a large-scale lightweight benchmark for 2d and 3d biomedical image classification},
  author={Yang, Jiancheng and Shi, Rui and Wei, Donglai and Liu, Zequan and Zhao, Lin and Ke, Bilian and Pfister, Hanspeter and Ni, Bingbing},
  journal={Scientific Data},
  volume={10},
  number={1},
  pages={41},
  year={2023},
  publisher={Nature Publishing Group UK London}
}

@article{dosovitskiy2020vit,
  title={An Image is Worth 16x16 Words: Transformers for Image Recognition at Scale},
  author={Dosovitskiy, Alexey and Beyer, Lucas and Kolesnikov, Alexander and Weissenborn, Dirk and Zhai, Xiaohua and Unterthiner, Thomas and  Dehghani, Mostafa and Minderer, Matthias and Heigold, Georg and Gelly, Sylvain and Uszkoreit, Jakob and Houlsby, Neil},
  journal={ICLR},
  year={2021}
}

@article{simeoni2025dinov3,
  title={Dinov3},
  author={Sim{\'e}oni, Oriane and Vo, Huy V and Seitzer, Maximilian and Baldassarre, Federico and Oquab, Maxime and Jose, Cijo and Khalidov, Vasil and Szafraniec, Marc and Yi, Seungeun and Ramamonjisoa, Micha{\"e}l and others},
  journal={arXiv preprint arXiv:2508.10104},
  year={2025}
}

@inproceedings{
teterwak2024large,
title={Is Large-scale Pretraining the Secret to Good Domain Generalization?},
author={Piotr Teterwak and Kuniaki Saito and Theodoros Tsiligkaridis and Bryan A. Plummer and Kate Saenko},
booktitle={The Thirteenth International Conference on Learning Representations},
year={2025},
}

@article{
zhang2024openood,
title={Open{OOD} v1.5: Enhanced Benchmark for Out-of-Distribution Detection},
author={Jingyang Zhang and Jingkang Yang and Pengyun Wang and Haoqi Wang and Yueqian Lin and Haoran Zhang and Yiyou Sun and Xuefeng Du and Yixuan Li and Ziwei Liu and Yiran Chen and Hai Li},
journal={Journal of Data-centric Machine Learning Research},
year={2024},
note={Dataset Certification}
}

@inproceedings{
hendrycks2017a,
title={A Baseline for Detecting Misclassified and Out-of-Distribution Examples in Neural Networks},
author={Dan Hendrycks and Kevin Gimpel},
booktitle={International Conference on Learning Representations},
year={2017},
}

@article{oquabdinov2,
  title={DINOv2: Learning Robust Visual Features without Supervision},
  author={Oquab, Maxime and Darcet, Timoth{\'e}e and Moutakanni, Th{\'e}o and Vo, Huy V and Szafraniec, Marc and Khalidov, Vasil and Fernandez, Pierre and HAZIZA, Daniel and Massa, Francisco and El-Nouby, Alaaeldin and others},
  journal={Transactions on Machine Learning Research}
}

@inproceedings{wang2022vim,
  title={Vim: Out-of-distribution with virtual-logit matching},
  author={Wang, Haoqi and Li, Zhizhong and Feng, Litong and Zhang, Wayne},
  booktitle={Proceedings of the IEEE/CVF conference on computer vision and pattern recognition},
  pages={4921--4930},
  year={2022}
}

@article{ndiour2020out,
  title={Out-of-distribution detection with subspace techniques and probabilistic modeling of features},
  author={Ndiour, Ibrahima and Ahuja, Nilesh and Tickoo, Omesh},
  journal={arXiv:2012.04250},
  year={2020}
}

@article{liu2025does,
  title={Does DINOv3 Set a New Medical Vision Standard?},
  author={Liu, Che and Chen, Yinda and Shi, Haoyuan and Lu, Jinpeng and Jian, Bailiang and Pan, Jiazhen and Cai, Linghan and Wang, Jiayi and Zhang, Yundi and Li, Jun and others},
  journal={arXiv preprint arXiv:2509.06467},
  year={2025}
}

@article{scholkopf1999support,
  title={Support vector method for novelty detection},
  author={Sch{\"o}lkopf, Bernhard and Williamson, Robert C and Smola, Alex and Shawe-Taylor, John and Platt, John},
  journal={Advances in neural information processing systems},
  volume={12},
  year={1999}
}

@article{aubreville2023comprehensive,
  title={A comprehensive multi-domain dataset for mitotic figure detection},
  author={Aubreville, Marc and Wilm, Frauke and Stathonikos, Nikolas and Breininger, Katharina and Donovan, Taryn A and Jabari, Samir and Veta, Mitko and Ganz, Jonathan and Ammeling, Jonas and van Diest, Paul J and others},
  journal={Scientific data},
  volume={10},
  number={1},
  pages={484},
  year={2023},
  publisher={Nature Publishing Group UK London}
}

@inproceedings{gutbrod2025openmibood,
  title={OpenMIBOOD: Open Medical Imaging Benchmarks for Out-Of-Distribution Detection},
  author={Gutbrod, Max and Rauber, David and Nunes, Danilo Weber and Palm, Christoph},
  booktitle={Proceedings of the Computer Vision and Pattern Recognition Conference},
  pages={25874--25886},
  year={2025}
}

@article{saikia2019comparative,
  title={Comparative assessment of CNN architectures for classification of breast FNAC images},
  author={Saikia, Amartya Ranjan and Bora, Kangkana and Mahanta, Lipi B and Das, Anup Kumar},
  journal={Tissue and Cell},
  volume={57},
  year={2019},
  publisher={Elsevier}
}

@inproceedings{amorim2020novel,
  title={A novel approach on segmentation of agnor-stained cytology images using deep learning},
  author={Amorim, Jo{\~a}o Gustavo Atkinson and Macarini, Luiz Antonio Buschetto and Matias, Andr{\'e} Vict{\'o}ria and Cerentini, Allan and Onofre, Fabiana Botelho De Miranda and Onofre, Alexandre Sherlley Casimiro and Von Wangenheim, Aldo},
  booktitle={2020 IEEE 33rd International Symposium on Computer-Based Medical Systems (CBMS)},
  year={2020},
  organization={IEEE}
}

@article{Wang2017ChestXRay8HC,
  title={ChestX-Ray8: Hospital-Scale Chest X-Ray Database and Benchmarks on Weakly-Supervised Classification and Localization of Common Thorax Diseases},
  author={Xiaosong Wang and Yifan Peng and Le Lu and Zhiyong Lu and Mohammadhadi Bagheri and Ronald M. Summers},
  journal={2017 IEEE Conference on Computer Vision and Pattern Recognition (CVPR)},
  year={2017},
  pages={3462-3471},
}

@article{tschandl2018ham10000,
  title={The HAM10000 dataset, a large collection of multi-source dermatoscopic images of common pigmented skin lesions},
  author={Tschandl, Philipp and Rosendahl, Cliff and Kittler, Harald},
  journal={Scientific data},
  volume={5},
  number={1},
  pages={1--9},
  year={2018},
  publisher={Nature Publishing Group}
}

\end{document}